\documentclass[review,number,sort&compress, 12pt]{article}

\pdfoutput=1

\usepackage{graphicx}
\usepackage{amssymb}
\usepackage{lineno}
\usepackage{listings}
\usepackage{color}

\usepackage{pdflscape}
\usepackage{afterpage}

\usepackage{subcaption}

\definecolor{lightgray}{rgb}{.98,.98,.98}
\definecolor{mediumgray}{rgb}{.6,.6,.6}
\definecolor{darkgray}{rgb}{.4,.4,.4}
\definecolor{purple}{rgb}{0.65, 0.12, 0.82}
\definecolor{darkgreen}{rgb}{0.008,0.617,0.067}

\definecolor{dkgreen}{rgb}{0,0.6,0}
\definecolor{gray}{rgb}{0.5,0.5,0.5}
\definecolor{mauve}{rgb}{0.58,0,0.82}
\definecolor{light-gray}{gray}{0.25}

\lstdefinelanguage{JSON}{
  keywords={:}
  keywordstyle=\color{darkgray},
  ndkeywords={class, export, boolean, throw, implements, import, this},
  ndkeywordstyle=\color{darkgray}\bfseries,
  identifierstyle=\color{black},
  sensitive=false,
  comment=[l]{//},
  morecomment=[s]{/*}{*/},
  commentstyle=\color{purple}\ttfamily,
  numberstyle=\color{black}\ttfamily,
  stringstyle=\color{blue}\textbf,
  morestring=[b]',
  morestring=[b]"
}

\lstdefinelanguage{JavaScript}{
  keywords={typeof, new, true, false, catch, function, return, null, catch, switch, var, if, in, while, do, else, 
case, break, emit},
  keywordstyle=\color{blue}\bfseries,
  ndkeywords={class, export, boolean, throw, implements, import, this, doc},
  ndkeywordstyle=\color{darkgreen}\bfseries,
  identifierstyle=\color{black},
  sensitive=false,
  comment=[l]{//},
  morecomment=[s]{/*}{*/},
  commentstyle=\color{purple}\ttfamily,
  stringstyle=\color{blue}\ttfamily,
  morestring=[b]',
  morestring=[b]"
}

\lstdefinelanguage{View}{
  keywords={:, function, emit, if, then}
  keywordstyle=\color{blue}\bfseries,
  ndkeywords={this, var, doc},
  ndkeywordstyle=\color{darkgreen}\bfseries,
  identifierstyle=\color{black},
  sensitive=false,
  comment=[l]{//},
  morecomment=[s]{/*}{*/},
  commentstyle=\color{purple}\ttfamily,
  numberstyle=\color{blue}\ttfamily,
  stringstyle=\color{blue}\textbf,
  morestring=[b]',
  morestring=[b]"
}

\lstdefinestyle{json}{
language=JSON,
backgroundcolor=\color{lightgray},
extendedchars=true,
basicstyle=\scriptsize\ttfamily,
showstringspaces=false,
showspaces=false,
numbers=none,
numberstyle=\scriptsize,
numbersep=9pt,
tabsize=2,
breaklines=true,
showtabs=false,
captionpos=b}

\lstdefinestyle{jscript}{
language=Javascript,
backgroundcolor=\color{lightgray},
extendedchars=true,
basicstyle=\scriptsize\ttfamily,
showstringspaces=false,
showspaces=false,
numbers=none,
numberstyle=\scriptsize,
numbersep=9pt,
tabsize=2,
breaklines=true,
showtabs=false,
captionpos=b}

\lstdefinestyle{java}{
  language=Java,
  aboveskip=3mm,
  belowskip=3mm,
  showstringspaces=false,
  columns=flexible,
  basicstyle={\footnotesize\ttfamily},
  numberstyle={\tiny},
  numbers=left,
  keywordstyle=\color{blue},
  commentstyle=\color{dkgreen},
  stringstyle=\color{mauve},
  breaklines=true,
  breakatwhitespace=true,
  tabsize=3
}

\lstdefinestyle{scala} {  
  morekeywords={ abstract,case,catch,
    char,class,
    def,else,extends,final,
    if,import,
    match,module,new,null,object,
    override,package,private,protected,
    public,return,super,this,throw,
    trait,try,type,val,var,with,implicit,
    macro,sealed
  },
  sensitive,
  morecomment=[l]//,
  morecomment=[s]{/*}{*/},
  morestring=[b]",
  morestring=[b]',
  aboveskip=3mm,
  belowskip=3mm,
  showstringspaces=false,
  columns=flexible,
  basicstyle={\footnotesize\ttfamily},
  numberstyle={\tiny},
  numbers=left,
  keywordstyle=\color{blue},
  commentstyle=\color{dkgreen},
  stringstyle=\color{mauve},
  breaklines=true,
  breakatwhitespace=true,
  tabsize=3
}

\lstdefinestyle{python}{
   language=Python,
   backgroundcolor=\color{lightgray},
   extendedchars=true,
   basicstyle=\scriptsize\ttfamily,
   showstringspaces=false,
   showspaces=false,
   numbers=none,
   numbersep=9pt,
   tabsize=2,
   breaklines=true,
   showtabs=false,
   captionpos=t,
   columns=flexible,
   numberstyle=\scriptsize\tiny\color{gray},
   keywordstyle=\color{red},
   commentstyle=\color{darkgreen},
   stringstyle=\color{blue},
   breakatwhitespace=true
}

\lstdefinestyle{Shell}{
language=csh,
backgroundcolor=\color{lightgray},
basicstyle=\scriptsize\ttfamily,
breaklines=true,
numberstyle=\scriptsize\tiny\color{gray}
}

\usepackage[figuresright]{rotating}

\graphicspath{
{figures/}
}

\usepackage[plainpages=false,pdfpagelabels]{hyperref}
\usepackage[capitalize]{cleveref}

\crefformat{figure}{#2Fig.\,#1#3}
\Crefformat{figure}{#2Figure #1#3}
\crefformat{equation}{#2Eq.\,#1#3}
\Crefformat{equation}{#2Equation #1#3}
\crefformat{section}{#2Sec.\,#1#3}
\Crefformat{section}{#2Section #1#3}

\usepackage{amsmath}
\usepackage{grffile}
\usepackage{multirow}
\usepackage{booktabs}

\usepackage[textwidth=18cm,textheight=24cm]{geometry}

\usepackage{xspace}

\usepackage{adjustbox}
\usepackage{array}

\newcolumntype{R}[2]{%
    >{\adjustbox{angle=#1,lap=\width-(#2)}\bgroup}%
    l%
    <{\egroup}%
}

\usepackage{biblatex}
\title{Classification of simulated radio signals using Wide Residual Networks for use in the search for extra-terrestrial intelligence.}

\author{G.\,A.\,Cox\thanks{IBM Watson Data Platform; adamcox@us.ibm.com} \and  S.\,Egly\thanks{Team Effsubsee} \and G.\,R.\,Harp\thanks{The SETI Institute, 189 Bernardo Ave., Mountain View, CA, 94043} \and J.\,Richards\footnotemark[3] \and S.\,Vinodababu\footnotemark[2] \and J.\,Voien\footnotemark[2]}

\date{}                                           

\begin{document}
\maketitle
\begin{abstract}
We describe a new approach and algorithm for the detection of artificial signals and 
their classification in the search for extraterrestrial intelligence (SETI). The 
characteristics of radio signals observed during SETI research are often most 
apparent when those signals are represented as spectrograms. Additionally, many 
observed signals tend to share the same characteristics, allowing for sorting of the 
signals into different classes.  For this work, complex-valued time-series data were 
simulated to produce a corpus of 140,000 signals from seven different signal classes. 
A wide residual neural network was then trained to classify these signal types using 
the gray-scale 2D spectrogram representation of those signals. An average $F_1$ score 
of 95.11\% was attained when tested on previously unobserved simulated 
signals. We also report on the performance of the model across 
a range of signal amplitudes.
\end{abstract}

\section{Deep Learning and SETI}


Advances over the last two decades in neural network training algorithms, increased computational power
and available data have had astonishing success with automatic image classification and similar applications. 
In this work, we apply these techniques to the unique case of signal classification of time-series radio signals. 



In a typical ETI search at radio frequencies, a radio telescope observes signals emanating 
from selected directions on the sky. After down-conversion and digitization, the raw data output of the 
telescope is a time series of digital voltage samples representing the electromagnetic field in the 
focal plane of the telescope. At the Allen Telescope Array (ATA), an array of 42, 6-meter dual-polarity offset-Gregorian
radio telescopes\cite{Welch2009}, a specialized program, called SonATA, 
sifts through these time series data looking for weak  
radio signals with telltale signs of 
artificial origin. At the heart of SonATA is a sensitive algorithm (Doubling Accumulation Drift Detector 
or DADD) which uses conventional digital signal processing techniques honed over decades of 
effort\cite{Cullers1985}.  DADD effectively detects just one kind of narrowband signal, a tone with a 
frequency that drifts linearly with time. While DADD has a low probability of generating false negatives, 
there are many different kinds of narrowband signals that generate false positives\cite{HarpRichardsTarterApJ2016}.
However, for a human, these signals are easily distinguishable from a drifting tone.  

In this paper, we implement convolutional neural networks (CNN) for the 
purposes of signal detection and classification, which may one day 
replace or complement the use of the DADD algorithm.
The CNN is a sensitive (low false negative rate) detector of several types of narrowband signals and 
additionally classifies those signals by type. The classifier makes our SETI search even more effective 
since signals generating false positive detections with DADD are correctly classified, effectively 
eliminating false positives without additional telescope time. Moreover, the CNN is a reliable detector 
of many different signal types (instead of just one). This expands our SETI search many-fold by 
generalizing the search to include a number of different signal types.


\subsection{The Spectrogram}


\begin{figure*}[t!]
    \centering
    \begin{subfigure}[t]{0.45\textwidth}
        \centering
        \includegraphics[height=1.775in]{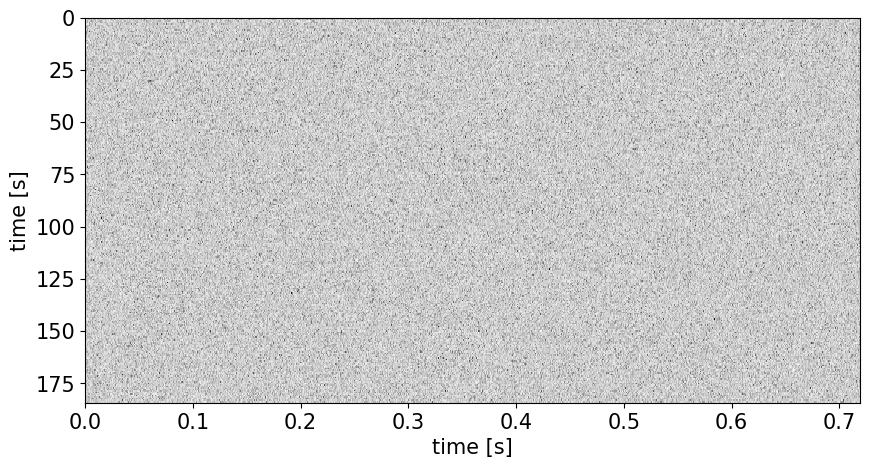}
        \caption{Observation in time-domain}
        \label{fig:setionata_timeseries}
    \end{subfigure}%
    ~ 
    \begin{subfigure}[t]{0.45\textwidth}
        \centering
        \includegraphics[height=1.775in]{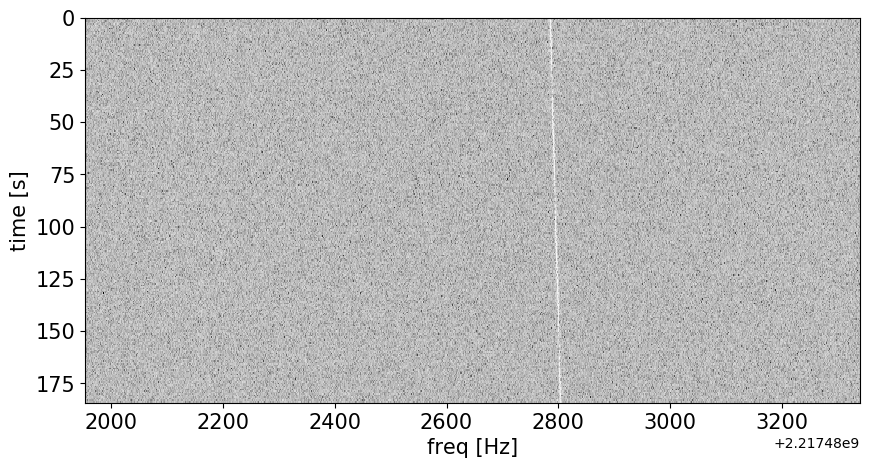}
        \caption{Observation as a spectrogram}
        \label{fig:setionata_spectrogram}
    \end{subfigure}
    \caption{Observation of a signal from the ISEE3 satellite on August 20, 2013. The raw data are shown in the time domain (a) and as a spectrogram (b).}
\end{figure*}

A common approach to signal detection is to transform the demodulated time-series 
signal to a spectrogram. The utility of a spectrogram
is highlighted with the following signal observed from the ISEE3 satellite on August 20, 2013.  
\Cref{fig:setionata_timeseries} shows the modulus of the time series dominated by noise and with a 
weak signal superimposed. The time series has been broken up (reshaped) into rows of length 1000 points and 
then arranged in time order from top to bottom. Visually, this representation of the time 
series is indistinguishable from pure noise. 

A spectrogram is a representation of a signal's power across the available frequency band, 
estimated by the squared modulus of the Fourier transform, as a function of time. 
To produce the spectrogram of the time series in \cref{fig:setionata_timeseries}, the data
in each row is passed through a frequency filter bank (based on the fast fourier transform) 
and the squared modulus of the resultant rows are arranged into a two dimensional array as 
before (\cref{fig:setionata_spectrogram}). Striking evidence of the superimposed signal is 
observed in the spectrogram. 

In both \cref{fig:setionata_timeseries} and \cref{fig:setionata_spectrogram}, the same 
data is presented as a 2D representation, suitable for input to a machine vision algorithm. 
The difference is that we have used our domain knowledge to choose the representation where the 
signal features are most easily detected.
The Fourier transformation efficiently adds structure that CNNs can utilize, while we do not know of any general purpose neural network architecture that can learn from the complex time-series with a reasonable quantity of sample data or within a reasonable training time.
Training on data similar to 
\cref{fig:setionata_spectrogram} will succeed in producing an effective data classifier 
with a modest training set (demonstrated below). 




It should be noted that deep neural networks have been used in other fields for the purpose of 
classifing spectrogram representations of time-series data; from music instrument 
recognition to noise detection at the 
gravitational-wave observatory, at the Laser Interferometer Gravitational-Wave 
Observatory (LIGO), and recently to discover new exoplanets with
Kepler Space Telescope data \cite{DBLP:journals/corr/ParkL15a, 2017arXiv170607446G, 2018ShallueVanderburgKepler90}.

\subsection{SETI Research with the Allen Telescope Array}

For full context of this work, we briefly describe normal signal detection operations at the ATA.
For twelve hours each day, the SETI Institute searches for persistent radio signals emanating from an extra-terrestrial source. 
The telescope array combines observations from up to 42 dishes into three, dual-polarization beams which can be pointed 
anywhere within the large ATA field of view (FOV, frequency dependent with a FWHM of 3.5${}^\circ$ at 1 GHz), 
in a process called \textit{beamforming}. 
These pencil-shaped synthetic beams measure only a small region of the sky at one time 
(FWHM 0.1${}^\circ$ at 1 GHz) centered on a target of interest. After downconversion and digital 
sampling, each beam produces voltage time series data at 104 million complex samples per second (MCSs), 
representing about 70 MHz of useful radio frequency bandwidth and center frequency tuned anywhere from 1-10 GHz.

The six data streams (three beams, two polarizations) are digitally processed as follows. A cascade of 
polyphase filter banks divide each high-speed beam stream into many parallel low-speed time series, each 
representing a sliver of about 8 kHz bandwidth. From here, each 
low-speed time series is examined independently for evidence of artificial signals. A spectrogram 
 is created and then searched with the DADD algorithm.

The full data acquisition system is managed by SonATA. It manages the nightly observations, 
controls beamforming, maintains and uses
an RFI database for comparison with observed signals, and performs signal classifications 
based on signal characteristics estimated from the 
DADD algorithm. SonATA also manages the follow-up observation procedure 
when signals are classified as \textit{candidate} signals.  A candidate signal means the data exhibits 
sufficient traits of a \texttt{narrowband} signal. 

After a candidate signal is found, SonATA performs a series of tests for persistence. 
In particular, SonATA directs the ATA to make observations
away from the direction from which the candidate signal appeared in order to determine that the 
candidate signal is \textit{not} observed elsewhere, and then reforms a beam toward the original location to 
affirm its continued presense. This process repeats up to five times, at which 
point a human is notified. These series of tests can 
take many tens of minutes to perform. 


In comparison to a neural net classifier, which we show can classify many different signal types at once, 
the DADD algorithm is limited. DADD can reliably find \texttt{narrowband} signals in spectrograms (low false 
negative rate) but can easily be fooled by signals that are clearly of another class (high false positive 
rate). Observing time is wasted every time DADD identifies a false positive, which must be followed up 
with additional observing. From this perspective, the neural net enabled 
classifier described here is more flexible and can potentially speed up the SETI search. 
On the other hand, a robust and efficient multi-class classifier opens up the possibility for SETI research
to consider different signal classes beyond \texttt{narrowband}. For example, a signal that is initially
detected as \texttt{narrowband} by the DADD may later possess a stochastically varying central frequency
or amplitude modulations that make DADD detection unreliable. A classifier trained on
signals with these different characteristics, however, would be more reliable. 

\subsection{Motiviation for Simulated Signals}

Recently, attemps have been made to cluster and classify candidate signals found by SonATA/DADD in 
the data set from 2013 to 2015, which contained slightly more than 4 million candidate signals. 
Promising approaches included one technique that utilized 
simple affine transformation followed by a comparator to examples of known signal types, 
and another technique which
utilized an autoencoder to extract a subset of features from spectrogram and then attempted to perform clustering
with the t-SNE alorithm \cite{Luus2018VAEForSETI}.  However, neither approach resulted in 
satisfactory classification capabilities. As such, 
this approach to develop a classifier based on a set of simulated data was conceived. 


\section{Simulations}

The simulated data set was designed to achieve a number of goals: 1) reproduce real SonATA data in basic 
structure, 2) simulate a feasible number of different signal classes, 3) test machine-learning model performance over a range of signal strengths, 4) produce signals with 
distinct characteristics between the classes, yet still produce signals that may be difficult to classify.
Ultimately, we simulated 140,000 time-series signals for the training
data. There were six signal classes, plus a set of simulations that contained only background noise. 

The data acquisition system at the ATA demodulates the observed signals from the GHz range and digitizes 
with 8-bit digitizers. This produces a complex-valued time-series data set with 8-bits for the real and imaginary
components. As such, our simulation program also outputs complex-valued 
(8-bit real, 8-bit imaginary) time-series data. It should be noted
that the "digitization rate" of the simulated data are unspecified, which means the time- and frequency-resolution
in the spectrograms is arbitrary. This has no effect, however, on the algorithm used to classify signals.

\subsection{Signal Classes and Characteristics}

\begin{figure*}[t!]
    \centering
    \begin{subfigure}[t]{0.49\textwidth}
        \centering
        \includegraphics[height=1.5in]{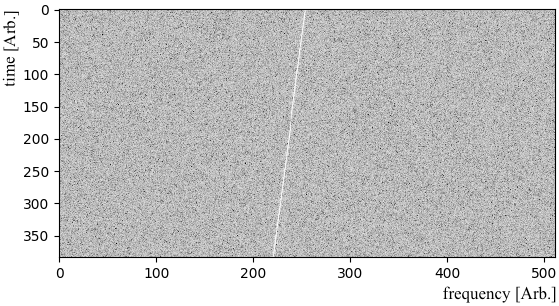}
        \caption{\texttt{narrowband}}
        \label{fig:sim_narrowband_example}
    \end{subfigure}
    \begin{subfigure}[t]{0.49\textwidth}
        \centering
        \includegraphics[height=1.5in]{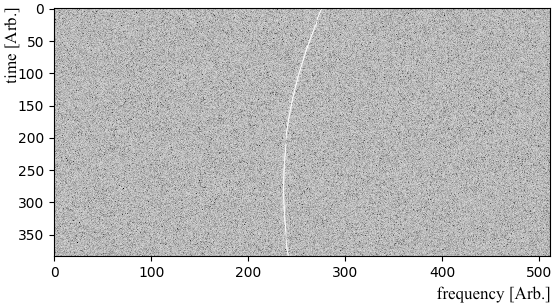}
        \caption{\texttt{narrowbanddrd}}
        \label{fig:sim_narrowbanddrd_example}
    \end{subfigure}
    \begin{subfigure}[t]{0.49\textwidth}
        \centering
        \includegraphics[height=1.5in]{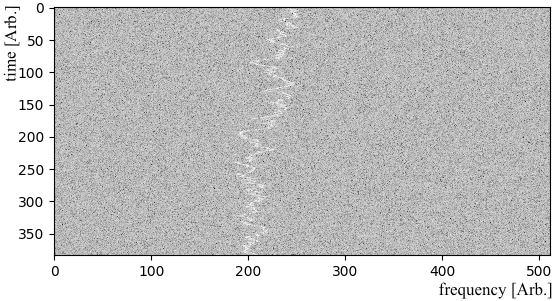}
        \caption{\texttt{squiggle}}
        \label{fig:sim_squiggle_example}
    \end{subfigure}
    \begin{subfigure}[t]{0.49\textwidth}
        \centering
        \includegraphics[height=1.5in]{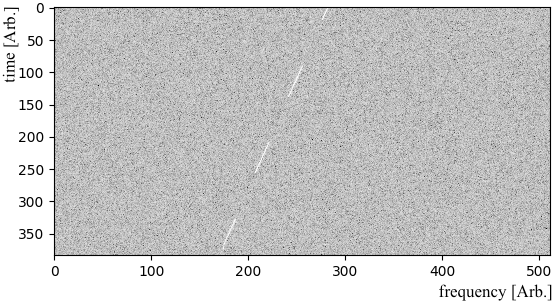}
        \caption{\texttt{squarepulsednarrowband}}
        \label{fig:sim_squarepulsednarrowband_example}
    \end{subfigure}
    \begin{subfigure}[t]{0.49\textwidth}
        \centering
        \includegraphics[height=1.5in]{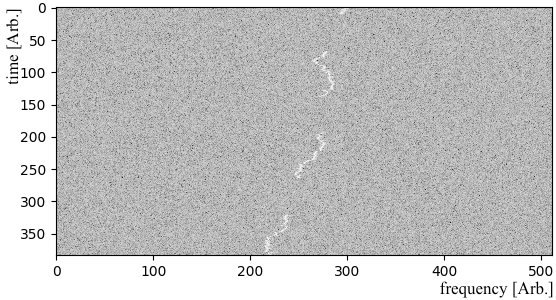}
        \caption{\texttt{squigglesquarepulsednarrowband}}
        \label{fig:sim_squigglesquarepulsednarrowband_example}
    \end{subfigure}
     \begin{subfigure}[t]{0.49\textwidth}
        \centering
        \includegraphics[height=1.5in]{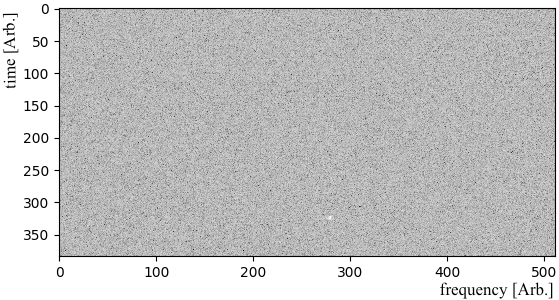}
        \caption{\texttt{brightpixel}}
        \label{fig:sim_brightpixel_example}
    \end{subfigure}
    \caption{Examples of the six simulated signal types. A seventh signal class was also simulated, 
    but contained only gaussian white noise and is not shown here.}
    \label{fig:sim_all_examples}
\end{figure*}

Based on domain knowledge from SETI Institute researchers, six signal classes were chosen to be simulated. These
classes represent, roughly, the more common signal types that have been observed. These signal types were
also relatively simple to simulate, making them attractive choices for this study. Signal types are specified by 
the apparant "shape" of the signal when observed as a gray-scale spectrogram. 
In the examples in this paper, white was chosen to 
represent the largest amplitude values and black 
are zero (\cref{fig:sim_all_examples}). The labels for the six simulated signal classes
were \texttt{brightpixel}, \texttt{narrowband}, \texttt{narrowbanddrd}, 
\texttt{squarepulsednarrowband}, \texttt{squiggle}, and \\
\texttt{squigglesquarepulsednarrowband}. The label
for the simulated data files that were noise only was \texttt{noise}. In the full training set there were 20,000 simulations for each of the seven classes. 

\subsubsection{Simulation Algorithm}

The simulated signals in the time-domain can all be described with a single equation,
\begin{equation} \label{eq:signal_basic} 
s(t)=A(t)\, \exp (i\omega(t)\,t+\phi )+n(t) 
\end{equation} 
where $A(t)$ is the signal amplitude, $\omega (t)$ is the frequency, and 
$\phi $ is a random phase offset. The noise component of all simulated signals, $n(t)$, was generated by random
sampling from a Gaussian with zero mean and width of $\sigma = 13.0$
for each of the real and imaginary components at every $t$. 
The seven signal types were generated using the following frequency and amplitude functions. The frequency function is
\begin{equation} \label{eq:signal_frequency} 
\omega(t)=\omega _{0} +(\omega _{1} + \dot{\omega}_{1} \, t) \, t+B\int _{0}^{t}\Omega (t)\, d\tau
\end{equation} 
where $\omega _{0}$ is the starting frequency,  $\omega _{1}$ is the drift rate, and $\dot{\omega}_{1}$ is its derivative. The $\Omega (t)$ term is a uniformly sampled random value between -1 and 1 that is updated at each
time step. The 
expression $\int _{0}^{t}\Omega (t)\, d\tau  $ corresponds to a random walk in frequency 
versus time and the arbitrary constant $B$ is referred to as the squiggle amplitude. 
The time-dependent amplitude is
\begin{equation} \label{eq:signal_amplitude} 
A(t) = A_o \cdot W( t \, | T, \, D, \, \phi_w)
\end{equation} 
where $W( t \, |T, \, D, \, \phi_w)$ is a square-wave modulation function that depends on a period, $T$, 
a duty cycle, $D$ and starting phase, $\phi_w$.

Each signal class was defined by a combination of setting some parameters to 0, and
allowing the others to be randomly sampled (\cref{tab:simparams}). For all signals types, 
the starting frequency, $\omega _{0}$, was sampled between [$-2\pi/3$, $2\pi/3$] and 
the drift rate, $\omega _{1}$, was sampled between [-7.324e-6, 7.324e-6].

The \texttt{narrowband} class (\cref{fig:sim_narrowband_example}) is the simplest signal type, defined by $\dot{\omega}_{1}=0$, $B=0$, and 
$W( t \, |T=L, \, D=1.0, \phi_w=0) = 1$, where $L$ is the full length of the simulation, which was fixed at $L=196608$. 
The equation for \texttt{narrowband} reduces to 

\begin{equation} \label{eq:narrow_band} 
s(t) = A_o\, \exp (i(\omega _{0} +(\omega _{1} \, t)\,t + \phi) + n(t)
\end{equation} 

The \texttt{narrowbanddrd} (DRD stands for ``drift rate derivative'') was defined by $\dot{\omega}_{1} \ne 0$, $B=0$, and $W( t \, |T=L, \, D=1.0, \phi_w=0) = 1$ (\cref{fig:sim_narrowbanddrd_example}). The \texttt{squiggle} class is defined by $\dot{\omega}_{1} = 0$, $B \ne 0$, and $W( t \, |T=L, \, D=1.0, \phi_w=0) = 1$ (\cref{fig:sim_squiggle_example}).

The \texttt{squarepulsednarrowband} (\cref{fig:sim_squarepulsednarrowband_example}), \texttt{squigglesquarepulsednarrowband} (\cref{fig:sim_squigglesquarepulsednarrowband_example}) \\
and \texttt{brightpixel} (\cref{fig:sim_brightpixel_example}) all 
have a time-dependent amplitude.  
For the \texttt{squarepulsednarrowband} and \texttt{squigglesquarepulsednarrowband}, the square-wave period, $T$, and duty cycle, $D$, were uniformly sampled
within the range specified in \cref{tab:simparams}. For the \texttt{brightpixel}, the period was fixed at $T=L$ and the duty cycle was 
uniformly sampled within a much smaller range, $D \in [0.0078125, 0.03125]$. The square-wave phase $\phi_w$, which sets the 
start-time of the modulation, was uniformily distributed between 7\% and 93\% of the full length of the signal, preventing the possibility
of the signal only being found near the very beginning or end of the simulation. 
Finally, signals of the \texttt{noise} class were defined by $A_o=0$.   

\begin{figure*}[t!]
    \centering
    \captionsetup{width=\textwidth}
    \begin{subfigure}[t]{0.49\textwidth}
        \centering
        \includegraphics[height=1.75in]{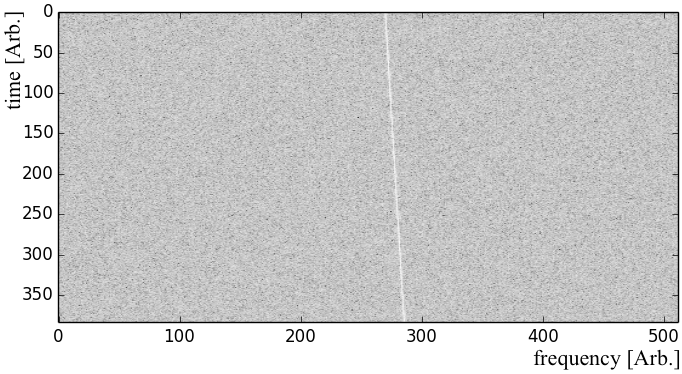}
        \caption{A moderately bright signal, with $A_o/13.0 = 0.25$}
        \label{fig:narrow_band_small_amp_visible}
    \end{subfigure}
     \begin{subfigure}[t]{0.49\textwidth}
        \centering
        \includegraphics[height=1.75in]{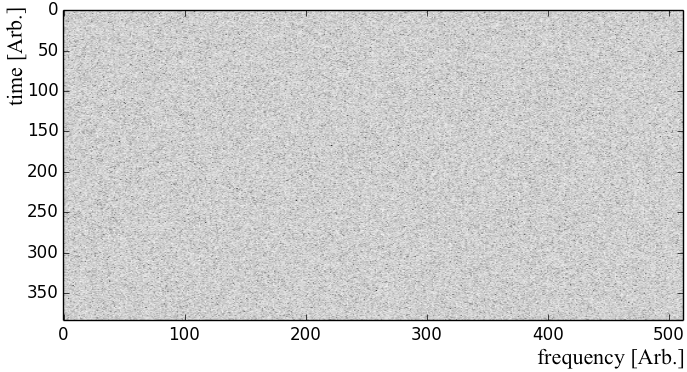}
        \caption{The dimmest possible signal, with $A_o/13.0 = 0.05$}
        \label{fig:narrow_band_small_amp_invisible}
    \end{subfigure}
    \caption{Two \texttt{narrowband} simulations at significantly different amplitudes, with the same drift 
    rate ${\omega_{1} = 0.001}$. The allowed range of $A_o$ for \texttt{narrowband} is found in \cref{tab:simparams}. The low-amplitude \texttt{narrowband} signal in \cref{fig:narrow_band_small_amp_invisible} is not visible by eye in this rendering.}
    \label{fig:narrow_band_small_amp}
\end{figure*}

\begin{figure*}[t!]
    \centering
    \begin{subfigure}[t]{0.49\textwidth}
        \centering
        \includegraphics[height=1.75in]{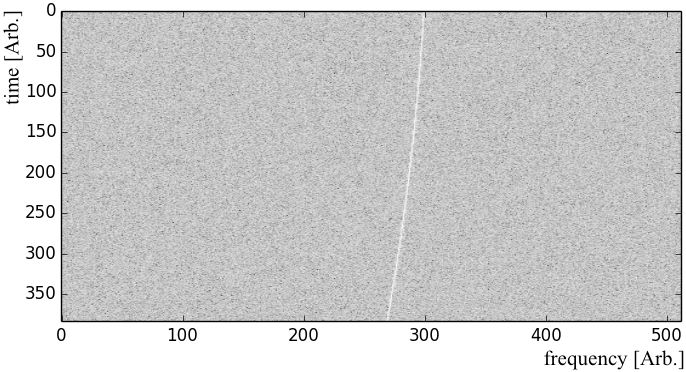}
        \caption{A \texttt{narrowbanddrd} with the smallest allowed ${\dot{\omega}_{1}}$.}
        \label{fig:narrow_band_drd_small_drd}
    \end{subfigure}
     \begin{subfigure}[t]{0.49\textwidth}
        \centering
        \includegraphics[height=1.75in]{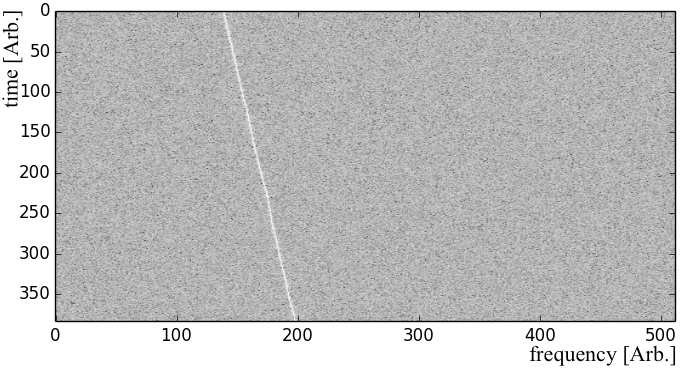}
        \caption{A \texttt{squiggle} with the smallest allowed $B$.}
        \label{fig:squiggle_small_amp_squiggle}
    \end{subfigure}
    \caption{A simulated \texttt{narrowbanddrd} with ${\dot{\omega}_{1} = -1.0 \cdot 10^{-8}}$ and initial drift rate 
    ${\omega_{1} = -9.542 \cdot 10^{-4}}$ and \texttt{squiggle} with ${B = 1.0 \cdot 10^{-4}}$ and initial drift rate
    ${\omega_{1} = 3.714 \cdot 10^{-3}}$.}
    \label{fig:narrow_band_drd_squiggle_compare}
\end{figure*}

The range of
values for the simulation parameters were tuned, by hand, to produce a 
set of simulated signals that appeared to match the real ATA data sets. They were also tuned in 
such a way to generate signals from one class that appeared very similar to signals from another. 
For example, \texttt{narrowband} signals with very small values of $A_o$ appear much 
like \texttt{noise} (\cref{fig:narrow_band_small_amp}), while
\texttt{narrowbanddrd} signals with small values of $\dot{\omega}_{1}$, and 
\texttt{squiggle} signals with small
values of $B$ appear very much like \texttt{narrowband} 
(\cref{fig:narrow_band_drd_small_drd} and \cref{fig:squiggle_small_amp_squiggle}, respectively). 

\afterpage{%
    \clearpage
    \thispagestyle{empty}
    \begin{landscape}
        \begin{table}[h!]
          \centering
            \resizebox{1.3\textwidth}{!}{%
            \begin{tabular}{l|lllllll}
                & $A_o$/13.0 & $\dot{\omega}_{1}$ ($10^{-8}$) & $B$ & $T/L$ & $D$\\
              \hline
              \texttt{narrowband} & [0.05, 0.4] &  0 & 0   & 1 & 1\\
              \texttt{narrowbanddrd} & [0.05, 0.4] &  [1, 8], [-1, -8] & 0 &  1 & 1\\
              \texttt{squiggle} & [0.1, 0.5] &  0 & [0.0001, 0.005]  & 1 & 1\\
              \texttt{squarepulsednarrowband} & [0.05, 0.4] & 0 & 0  & [0.15625, 0.46875] & [0.05, 0.9]\\
              \texttt{squigglesquarepulsednarrowband} & [0.1, 0.5] & 0 & [0.0001, 0.005] & [0.15625, 0.46875] & [0.15, 0.8]\\
              \texttt{brightpixel} & [0.05, 0.75] &  0 & 0 & 1 & [0.0078125, 0.03125]\\
              \texttt{noise} & 0 & - & -  & - & - \\
            \end{tabular}}
          \caption{Values for simulation parameters. For each simulation, values were uniformly sampled within the range or set to the value specified. For all signals, $\omega _{0} \in [-2\pi/3, 2\pi/3]$ and $\omega _{1} \in [-7.324, 7.324]\cdot 10^{-6}$}
          \label{tab:simparams}
        \end{table} 
    \end{landscape}
    \clearpage
}

As may be noticed by those with signal processing experience, this model would generate spectrograms that would alias
high-frequency signals to low (negative) frequency once $\omega(t) + \phi > \pi$.  Similarly, low-frequency signals would 
alias up to high-frequencies in the spectrogram once $\omega(t) + \phi < -\pi$. To prevent this aliasing effect, code was included in the simulation that set $A_o=0$ when the frequency crossed either of these thresholds. 

\section{Neural Network Signal Classifier}
We frame the challenge of classifying radio signal time series of complex
amplitudes as an image recognition task on their 2D spectrograms and apply
Wide Residual Network with 34 convolutional layers and a widening factor of 2.

\begin{figure*}[h!]
    \centering
    \includegraphics[height=3.1in]{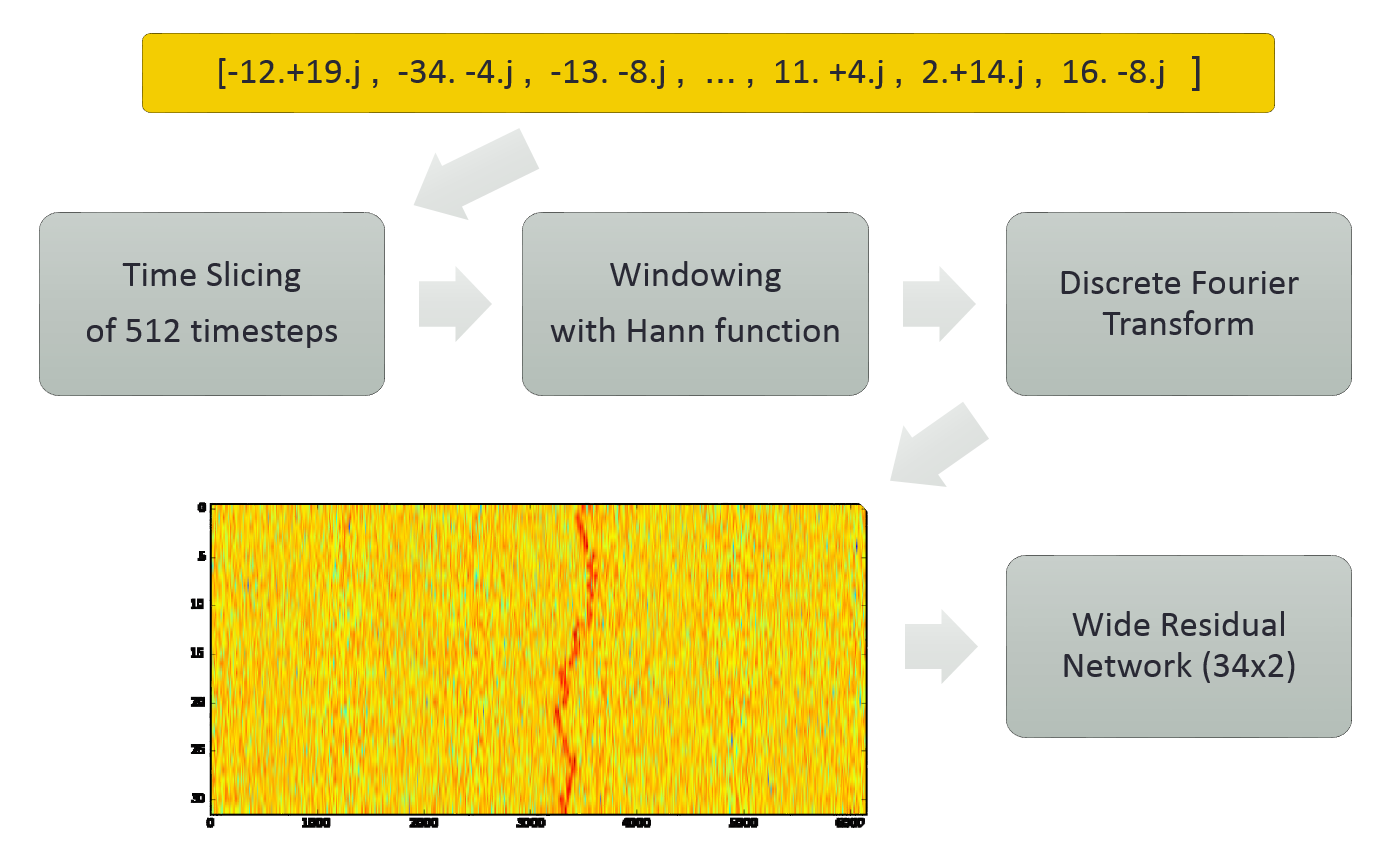}
    \caption{Classification Pipeline}
    \label{fig:approach_overview}
\end{figure*}

\subsection{Recasting as Image Classification}
The motivation to recast signal classification as an image classification 
problem comes from the fact that humans regularly classify these signals by 
inspecting a spectrogram.  For this image recognition task, a spectrogram for a 
given time series of complex amplitude is created by first reshaping the data to
height and width more appropriate for our neural network classifier.  Then a Hanning 
window and Fourier transform are applied.  The square of the absolute value of 
the result provides the spectrogram, which we take the logarithm of to produce 
one of the images to be supplied to our convolutional neural network.  The 
second is the phase of the Fourier transformed signal, which was included for
the purpose of providing additional useful features to the neural network
models tested. Although a followup investigation indicated it did not increase
the classification accuracy of our final model trained on simulated signals,
phase may be useful when using real signals that have
properties not included in our simulations, such as polarization, and is an
area of future research.

The reciprocally dependent time- and frequency-resolution of a spectrogram can be arbitrarily
chosen within the bounds of the total length of the signal. This choice results
in spectogram images of different shapes (height and width). 
While these 
differences can change signal detectability as a human perceives it, for the
purposes of classification by a convolutional neural network, we choose 
an aspect ratio close to square, since downsampling will reduce both
horizontal and vertical resolution significantly, and a reduction to near zero
would be unproductive.

\subsection{Exploration of CNN Architectures}

Deep learning is an active area of research in which the advances of deep neural
networks have produced near-human or superhuman accuracy in supervised machine 
learning tasks, including image recognition tasks \cite{AlexNet}.
Convolutional neural networks (aka convnets or CNNs) are at the core of these
advances, and the gains in state-of-the-art performance demonstrated by ResNet
is a go-to example\cite{ResNet}.  To accurately classify these images
generated from radio signal time series, we test multiple convolutional neural
network architectures in the classification task using only the spectrogram and
phase images as input features.

The milestone advance in image classification performance seen in the 2015 
ImageNet showing by the ResNet team suggests that this tried-and-true network 
is a suitable baseline for conducting these experiments.  Augmentations and 
further improvements to the ResNet architecture have been presented since, so 
we tested these in a benchmark against ResNet and each other to determine which 
performed best on this particular dataset.  Although these competing networks 
were able to show state-of-the-art performance on various tasks and datasets, 
their performances are close enough that for this task, somewhat different from 
the typical benchmarks for image recognition, empirical evidence is sought to 
determine which network best classifies simulated radio signal spectrograms.

\subsubsection{Residual Network (ResNet)} 
ResNet introduced the residual connection between convolutional layers in a 
very deep convolutional neural network in order to combat the loss of signal 
backpropagation.  Previously, very deep networks tended not to train 
successfully, but this skip connection that bypasses a layer's nonlinearity 
allows gradients to backpropagate further in the network and allow for deeper, 
more expressive networks.  Hence, we experimented with ResNets up to the limit 
of computational constraints for these medium-resolution images.

\subsubsection{Wide Residual Network (WRN)}
The development of wide residual networks was based on the observations that 
increasing the depth of ResNet provided diminishing returns on network 
performance improvement, and offered shallower convolutional networks with 
more convolutional filters at each layer as a solution\cite{WideResNet}.  
Additionally, the authors change the order of convolution, batch normalization, 
and activation, and add dropout to tune their architecture to train faster and 
perform competitively with deeper ResNet models.  In our experiments, we used 
the variations they had the most success with; namely, using the same 3x3 basic 
convolutional block (as opposed to a bottleneck block), and a dropout rate of 
0.3.  

\subsubsection{Densely Connected Residual Network (DenseNet)}
DenseNet extends the idea of residual connections introduced by ResNet by 
adding residual connections not only between consecutive convolutional layers, 
but also between all subsequent convolutional layers\cite{DenseNet}.  The dense 
residual block allows the gradient signal to skip more layers, and which more 
closely ties the loss function to earlier layers of the network.  The 
additional skip connections also are thought to encourage feature reuse by 
sending signal from multiple convolutional layers to later layers, leading to 
more expressive power in a more compact network.  

\subsubsection{Dual Path Network (DPN)}
Finally, we experimented with dual path networks, which integrate both residual 
networks and densely connected residual networks to realize the advantages of 
each, while sharing weights to maintain a reasonable model 
complexity\cite{DualPathNetworks}.  

\subsection{Comparison of Models}
We trained several models for the ResNet, WRN, DenseNet, and DPN architectures,
since each architecture can be implemented with varying sizes.  The accuracy of
a learned classifier depends on this size because less complex networks cannot
necessarily express as complex a pattern as a larger network, but the larger
networks do not necessarily succeed in learning the patterns they can express.
\Cref{tab:model_comparison} lists the models trained for each architecture type,
with the best performing model shown in bold.

\begin{table}[h!]
  \centering
    \begin{tabular}{l|lll}
       Architecture  & Models Tested & Accuracy & Parameters \\
      \hline
      ResNet & ResNet-18, ResNet-50, \textbf{ResNet-101} & 94.99 & 42.6M   \\
      & ResNet-152, and ResNet-203 & &  \\
      Wide ResNet & \textbf{WRN-34-2}, WRN-16-8, & 95.77 & \ 1.9M   \\
      & and WRN-28-10 &  &    \\
      DenseNet & DenseNet-161, \textbf{DenseNet-201} & 94.80 & 18.3M   \\
      Dual Path Network & \textbf{DPN-92}, DPN-98, and DPN-131 & 95.08 & 35.1M   \\
    \end{tabular}
  \caption{Comparison of model architectures. Accuracy is measured with a cross-validation subset of the full training set. The number of parameters are given for the best-performing model (bold) of each type.}
  \label{tab:model_comparison}
\end{table}


\begin{figure*}[h!]
    \centering
        \centering
        \includegraphics[height=3.1in]{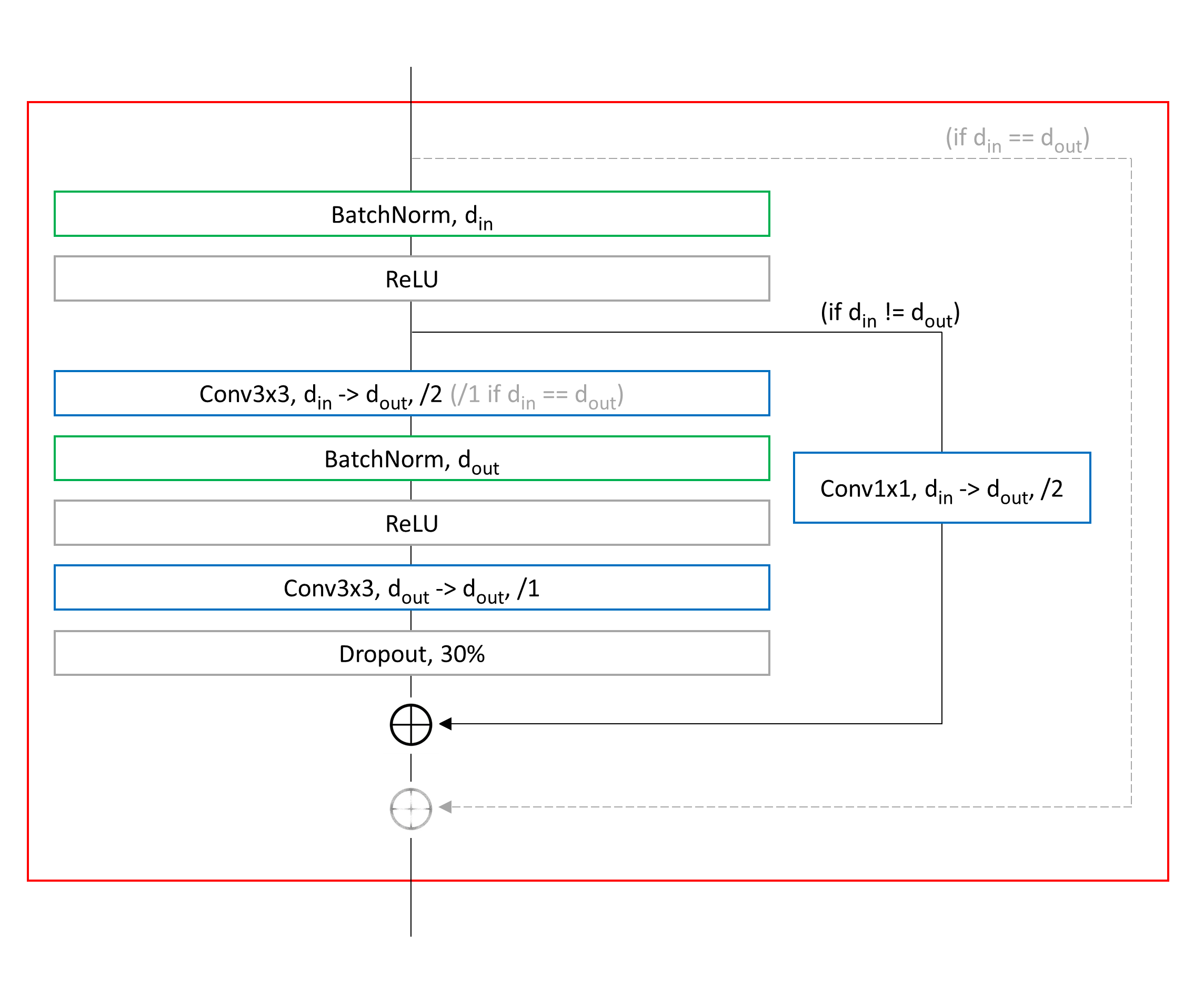}
        \caption{Basic Block: two 3x3 2D convolutions with batch normalization, ReLU nonlinearity, and a residual connection}
        \label{fig:model_block}
\end{figure*}

\begin{figure*}[h!]
        \centering
        \includegraphics[height=7.5in]{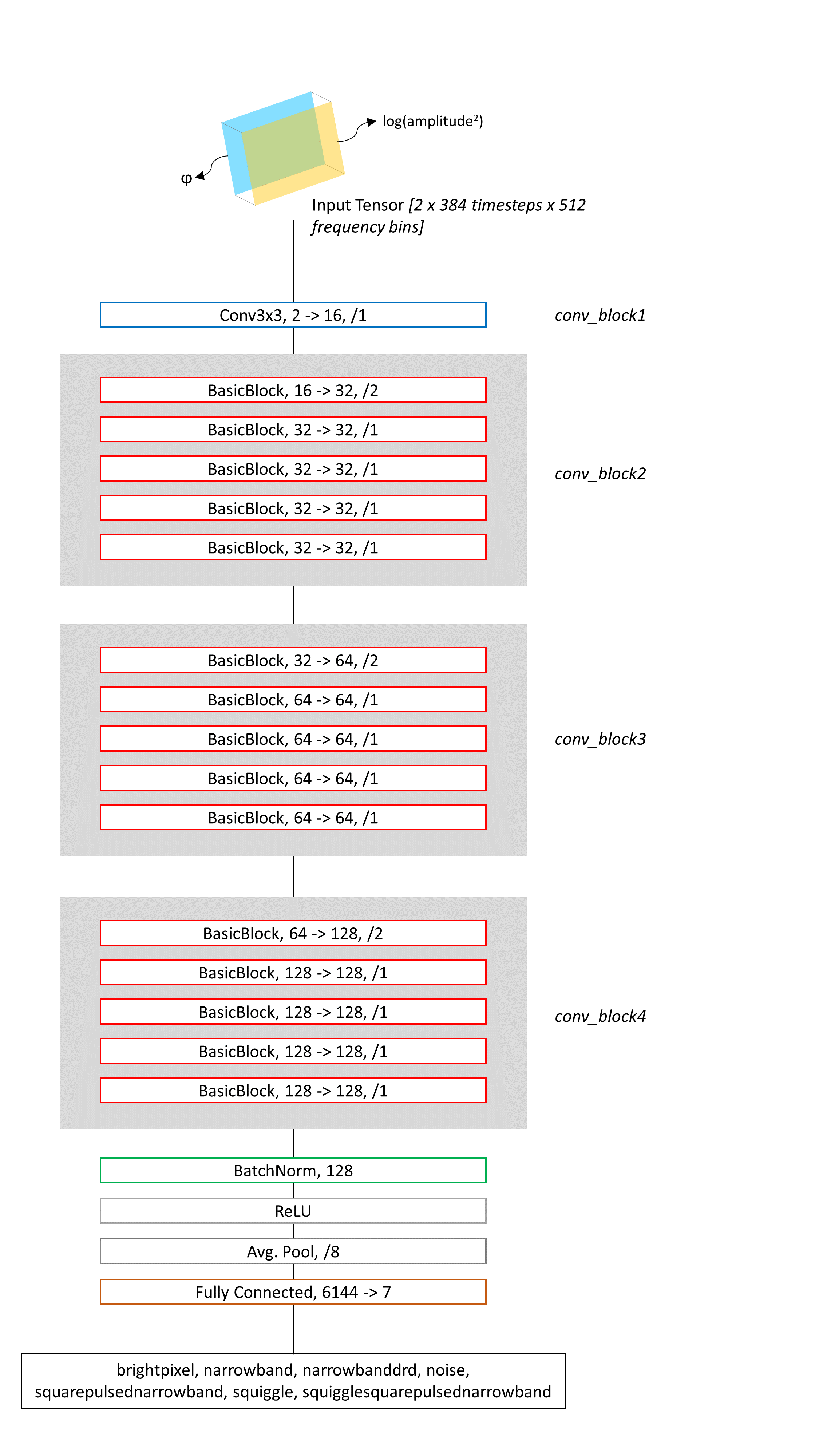}
        \caption{Complete WRN-34-2: a series of Basic Blocks grouped into three sections of increasing number of features planes and decreasing image resolution}
        \label{fig:model_network}
    \label{fig:model_architecture}
\end{figure*}

The accuracies included in \cref{tab:model_comparison} are validation accuracies on the same
train-validate split for all architectures. The performances of the four best
models are comparable, with the WRN-34-2 (\cref{fig:model_architecture})
only slightly outperforming the rest.
However, a very significant basis for choosing the WRN-34-2 over the
other models is the size of the model. The WRN-34-2 contains only 1.9M
parameters (also called weights). In comparison, the second smallest model, the
DenseNet-201, is nearly ten times as large. The ResNet-101 is the largest, with
over twenty times as many parameters. For the purposes of quick training and
inference on-site at the ATA, it is ideal to have a small memory footprint.

\subsection{Wide Resnet Ensemble}

The comparison of these four network architectures showed similar and
encouraging results, but the edge that WRN-34-2 had over the others led
us to continue working to improve its performance. A common practice to increase
generalizability of a machine learning model that threatens to overfit on
training data is to create an ensemble of models. For this task, ensemble
averaging was used. Five WRN-34-2 models were trained on different
training data but with the same hyperparameters and training strategy. While
the comparative study of architectures used the same four of five folds for
training and the fifth for validation, the ensemble members trained on the five
distinct four-fold subsets of data, as with k-fold validation for $k = 5$.

To evaluate the ensemble model, each of the five member models outputs its
softmax predictions and the average of these scores is taken as the final score.
With no validation data left over, this ensemble model was evaluated on new test
data provided as part of the code challenge, which yielded an accuracy of 94.99\%.

\section{Final Model Performance}

\subsection{Classification Report}

The model with the best validation set accuracy, the WRN-34-2 using a 5-fold averaging, was then 
tested using a separate test set withheld during the training phase of all models. 
As this work was performed in the context of an online code challenge, the other models
described in the previous section were not tested with this test data set. 

\Cref{tab:classification_accuracy} and \cref{tab:confusion_matrix} display the performance
of the model on the test set of 2496 simulations.

\begin{table}[h!]
  \centering
    \begin{tabular}{l|llll}
        & $N$ & precision & recall & $F_1$ \\
      \hline
      \texttt{brightpixel} &  385 & 0.991  &  0.857   & 0.919    \\
      \texttt{narrowband} & 355 & 0.994 &  0.944  & 0.968       \\
      \texttt{narrowbanddrd} &  348 & 0.969 &  0.977  & 0.973      \\
      \texttt{noise} &  368  & 0.785  & 0.995  &  0.877    \\
      \texttt{squarepulsednarrowband} &   385 &  0.975  &  0.925   & 0.949     \\
      \texttt{squiggle} & 322 & 1.000  & 0.997 &  0.998       \\
      \texttt{squigglesquarepulsednarrowband} &   332  &  1.000  &   0.970  &  0.984     \\
    \end{tabular}
  \caption{Model Performance Scores. The number of simulations of each class in the test set is given by $N$.}
  \label{tab:classification_accuracy}
\end{table} 

\begin{table}[h!]
  \centering
    \begin{tabular}{l|lllllll}
        Actual / Predicted & \texttt{bp} & \texttt{nb} & \texttt{drd} &  \texttt{no}  & \texttt{sqnb} & \texttt{sgl} & \texttt{sglsqnb}   \\
      \hline
      \texttt{bp} & 330 & 0 & 0 &  55 & 0 & 0 & 0 \\
      \texttt{nb} &  0  & 335 & 10 & 5 & 5 & 0 & 0   \\
      \texttt{drd} &   0  &  0 & 340  &  7  &  1  &  0  &  0  \\
      \texttt{no} &   1  & 0 &  0  & 366  & 1  & 0 &  0\\
      \texttt{sqPnb} &   2  &  2  &  1  & 24 &  356 &  0  &  0 \\
      \texttt{sgl} &   0  &  0 &  0  & 1 &  0 & 321 &  0   \\
      \texttt{sglsqPnb} &  0  & 0  &  0  &  8  &  2  &  0  & 322 \\
    \end{tabular}
  \caption{Confusion Matrix. Actual classifications are along each row and Predicted classification counts are along the columns. For example, there were 322 \texttt{squiggle} signals. Of those, 321 were predicted to be \texttt{squiggle}s and one was predicted to be \texttt{noise}. 
  Labels have been abbreviated: \texttt{bp} for \texttt{brightpixel}, \texttt{nb} for \texttt{narrowband}, \texttt{drd} for \texttt{narrowbanddrd}, \texttt{no} for \texttt{noise}, \texttt{sqPnb} for \texttt{squarepulsednarrowband}, \texttt{sgl} for \texttt{squiggle}, and \texttt{sglsqPnb} for \texttt{squigglesquarepulsednarrowband}.}
  \label{tab:confusion_matrix}
\end{table} 

Clearly the largest source of uncertainty was distinguishing \texttt{brightpixel} signals from \texttt{noise}. 
As we'll see in the next section, this was due to very low-amplitude \texttt{brightpixel} 
signals, as one would intuitively expect. 

\subsection{Model Performance Characteristics}\label{sec:mod_perf_char}

In order to briefly explore the performance characteristics of this trained model in a controlled way, 
14 new sets of test data were generated, each with 250 signals of each class. For each set, 
the signals were simulated with a fixed signal amplitude, $A/13.0 =$ 0.008, 0.01, 0.02, 0.04, 0.05, 0.06, 0.07, 0.08, 0.09, 0.1, 0.12, 0.16, 0.2, or 0.4.
Many of these signal ampliutdes, it should be noted, are below the amplitudes of signals found in 
the \textit{training} data set, allowing us to explore the model performance slightly outside of 
range of signal amplitudes on which it was trained. 
All other parameters of the test data, however, remained consistent with the training data.

\begin{figure*}[t!]
    \centering
    \begin{subfigure}[t]{0.5\textwidth}
        \centering
        \includegraphics[height=1.8in]{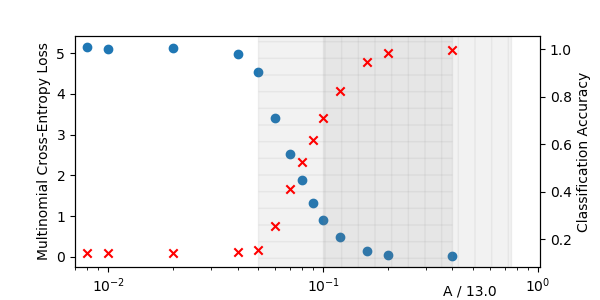}
        \caption{}
        \label{fig:model_char_logloss_and_acc_v_snr}
    \end{subfigure}%
    ~ 
    \begin{subfigure}[t]{0.5\textwidth}
        \centering
        \includegraphics[height=1.8in]{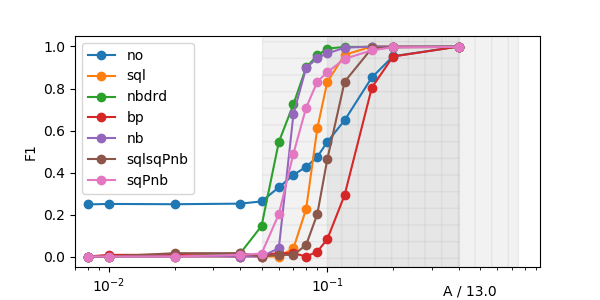}
        \caption{}
        \label{fig:model_char_f1_v_snr}
    \end{subfigure}
    \caption{Multinomial cross-entropy loss scores, classification accuracy, and $F_1$ scores versus the signal 
    amplitude ($A/13.0$). The horizontally hatched region represents the signal amplitudes in the training data
    with amplitudes in the range [0.05, 0.4], while the vertically hatched region spans the amplitudes 
    from [0.1, 0.75]. Refer to \cref{tab:simparams} for the signal classes that were simulated in these regions.}
\end{figure*}

For each of these test sets, we performed inference and recorded the model's multinomial cross-entropy loss, 
classification accuracy (\cref{fig:model_char_logloss_and_acc_v_snr}), 
and $F_1$ score (\cref{fig:model_char_f1_v_snr}). 
The model performs as expected. Signals with smaller amplitudes were more difficult to  
classify and tend to be classified as \texttt{noise}. Also, the model does not appear to have any classification
power outside of the trained signal amplitude space. 

Furthermore, the onset rise of each class's $F_1$ score (\cref{fig:model_char_f1_v_snr}) can be intuitively explained
by considering the average amount of power per pixel in each signal, akin to the apparent 
\textit{brightness} of the signal in the spectrogram. The \texttt{brightpixel} signals have 
lowest total power, overall. These signals have zero amplitude for most of the simulation except for a 
brief moment at a particular frequency. As such, the model struggles to recognize these signals the most. 
The simulations that contain a non-zero 
$B$ (\texttt{squiggle} and \texttt{squigglesquarepulsednarrowband}) also have
a reduced apparent brightness in the spectrogram. The stochastic flucutations of these signals result in
the power being 
spread across a larger bandwidth of frequencies during any particular time range of the 
simulation, as compared with \texttt{narrowband}. As such, one expects the classification and recall 
of those signals to be reduced relative to the other signals that appear brighter in the spectrogram. 
The \texttt{narrowband} and \texttt{narrowbanddrd} have the earliest onset of significant $F_1$ scores, followed
by \texttt{squarepulsednarrowband}, consistent with the argument that the brighter signal types are more
easily recognizable.

\section{Conclusions}

This work demonstrates the potential usefulness of applying contemporary convolutional neural networks
to SETI research. With further improvements, these algorithms may soon become part of 
SETI research data acquisition systems.

Subsequent work building on the results here could focus on a number of aspects.  Clearly, the 
precision and recall scores for the different signal classes were quite good. The most
immediate next step would be to build new models trained on data generated using a 
larger range of simulation parameters for each class (\cref{tab:simparams}). 
Additionally, more signal types could be added to training and test sets. For example, a 
common signal type not included in our repertoire is a short burst of power over all 
frequencies sometimes identified with radar pulses. 

The model's performance falloff just at the lower-bound of the signal amplitudes found 
in the training data 
 (\cref{sec:mod_perf_char}) leads to a particular question: to what small signal amplitude
can we train a model of the same network architecutre and still retain robust classification accuracy? 
That is, if we were to construct new training data with smaller signal amplitudes and retrain new models, how
small in ampliutude can we go before the models fail to accuractly classify signals?

Besides new signal types and characeteristics, the noise component of the signal, 
$n(t)$, could be more realistic. For 
this study, we used a very simple gaussian white noise model. 
In a previous version of the training data set, however,
we  used real observations from the Sun as the background component, $n(t)$. 
Observations from the Sun do not 
have constant power at all frequencies and are non-stationary. 
This background data was not used in this work, but may be the basis for simulations in future work. 

Real observations from the ATA do not always contain just one signal or signal type in the spectrogram. Therefore, 
another future work would be to include a more complete set of commonly-observed signal types within the same simulation and build
models that can find multiple signal types with techniques similar to those used to perform object detection
in ordinary daily human photographs. 

The set of simulated data could also be improved through the use of some type of generative network. This would require
some effort, though not unreasonably burdensome, to hand-label real spectrogram observations. An autoencoder
and t-SNE clustering approach was already taken with ATA data from 2013 to 2015 to 
cluster signals \cite{Luus2018VAEForSETI}. 
Although these simulations from generative network model would appear to be more realistic, the signals would not have 
controllable (or learnable) parameters such as $A_o$, $B$, $\dot{\omega}_{1}$, etc.

Finally, the DADD algorithm currently in use at the ATA estimates a linear fit to  
observed signals, with parameters including the signal power,
initial signal frequency, $\omega _{0}$, and drift rate, $\omega_{1}$. 
This regression is important because it allows for prediction of the future signal frequency, 
which is needed to identify the same signal in a subsequent observation. 
We speculate that either the parameters of human-constructed models (Eqs.\,\ref{eq:signal_basic} - \ref{eq:signal_amplitude}) for each signal class could be 
estimated through regression after signals classification with the 2D CNN, or that some
type of recurrent neural network, or causal convolutional architecture would be able to estimate
signal characteristics and predict future signal frequency by using each row of the spectrogram 
as the input vector at each time-step. 



In conclusion, we report here the adaptation of a convolutional neural network, specifically a wide residual network, to 
the problem of signal discovery and classification relevant to SETI. We find that by treating 
spectrograms as if they were images, we can train an image classifying network and achieve very good 
results. The sensitivity of the neural network detector is comparable to that of a finely honed 
conventional processing algorithm, and that the signal classification accuracy is quite good. Using 
such signal classifications will improve the efficiency of radio SETI, by providing extra information 
that can be used to decide whether to follow-up on detected signals. Compared to the conventional 
algorithm, the neural network approach lacks the parameterization of signals necessary for 
extrapolation to later times. We suggest that this limitation may be overcome with a traditional approach or with additional neural networks.

\section{Acknowledgements}

We'd like to thank Galvanize, Skymind, Nimbix, and The SETI League for their financial contribution and 
to the hard work put in by many employees of those organizations that ensured a successful code 
challenge. Thanks to IBM for providing significant compute and data storage.
We acknowledge the helpful suggestions from Francois Luus of IBM Research South Africa. 
Special thanks to Graham Mackintosh for establishing the IBM-SETI Institute partnership.


\printbibliography

\end{document}